\pdfoutput=1

\documentclass[11pt]{article}

\usepackage[preprint]{acl}

\usepackage{times}
\usepackage{latexsym}
\usepackage{bm}
\usepackage[T1]{fontenc}

\usepackage[utf8]{inputenc}

\usepackage{microtype}

\usepackage{inconsolata}

\usepackage{graphicx}

\usepackage{booktabs}
\usepackage{caption}
\usepackage{fontawesome5}
\usepackage{makecell}
\usepackage{multirow}
\usepackage{tcolorbox}
\usepackage{colortbl}
\usepackage{color}
\usepackage{bbding}
\usepackage{pifont}
\usepackage{mdframed}
\definecolor{Black}{rgb}{0.0, 0.0, 0.0}
\title{Can Pruning Improve Reasoning? Revisiting Long-CoT Compression with Capability in Mind for Better Reasoning} 

\author{
 \textbf{Shangziqi Zhao\textsuperscript{1}}\thanks{These authors contributed equally as co-first authors.}, 
 \textbf{Jiahao Yuan\textsuperscript{2,5,6}}\footnotemark[1]\thanks{Project Lead.},
  \textbf{Jinyang Wu\textsuperscript{3}\footnotemark[1]},
  \textbf{Zhenglin Wang\textsuperscript{4}} \\
 \textbf{Guisong Yang\textsuperscript{5}}\thanks{Corresponding Authors.}, 
  \textbf{Usman Naseem\textsuperscript{6}}\footnotemark[3]
\\
\\
 \textsuperscript{1}XJTU,  \quad
  \textsuperscript{2}ECNU, \quad
   \textsuperscript{3}THU,  \quad
    \textsuperscript{4}SEU,  \quad
 \textsuperscript{5}USST,  \quad
 \textsuperscript{6}MQ
\\
 \href{mailto:zszq618@stu.xjtu.edu}{\texttt{zszq618@stu.xjtu.edu}}, \quad
  \href{mailto:jhyuan.cs@gmail.com}{\texttt{jhyuan.cs@gmail.com}}, \quad
  \href{mailto:wu-jy23@mails.tsinghua.edu.cn}{\texttt{wu-jy23@mails.tsinghua.edu.cn}}
  \\
    \href{mailto:zhenglin@seu.edu.cn}{\texttt{zhenglin@seu.edu.cn}}, \quad
  \href{mailto:gsyang@usst.edu.cn}{gsyang@usst.edu.cn}, \quad
  \href{mailto:usman.naseem@mq.edu.au}{usman.naseem@mq.edu.au}
}

\usepackage{float}
\usepackage{tabularx}
\usepackage{booktabs}
\usepackage{amsthm,amsmath,amssymb}
\usepackage{mathrsfs}
\usepackage{algorithm}
\usepackage{algorithmic}
\usepackage{amsmath}

\usepackage{enumitem}
\usepackage{longtable}
\begin{document}
\maketitle

\begin{abstract}
Long chain-of-thought (Long-CoT) reasoning improves accuracy in LLMs, yet its verbose, self-reflective style often hinders effective distillation into small language models (SLMs). We revisit Long-CoT compression through the lens of capability alignment and ask: \textit{Can pruning improve reasoning?} We propose \textbf{Prune-on-Logic}, a structure-aware framework that transforms Long-CoT into logic graphs and selectively prunes low-utility reasoning steps under self-verification constraints. Through systematic analysis across three pruning strategies—targeting entire chains, core reasoning, and verification, we find that verification pruning consistently improves accuracy while reducing token usage, whereas pruning reasoning steps or indiscriminate pruning degrades performance. Our study reveals that effective pruning aligns supervision with model capacity rather than merely shortening inputs. Gains hold across tasks, model scales, and CoT capability, with larger models benefiting more from pruning due to richer but more redundant reasoning. Our empirical findings highlight pruning as a structural optimization strategy for aligning CoT reasoning with SLM capacity. 
\end{abstract}
\begin{figure}[t]
\centering
\includegraphics[width=\linewidth]{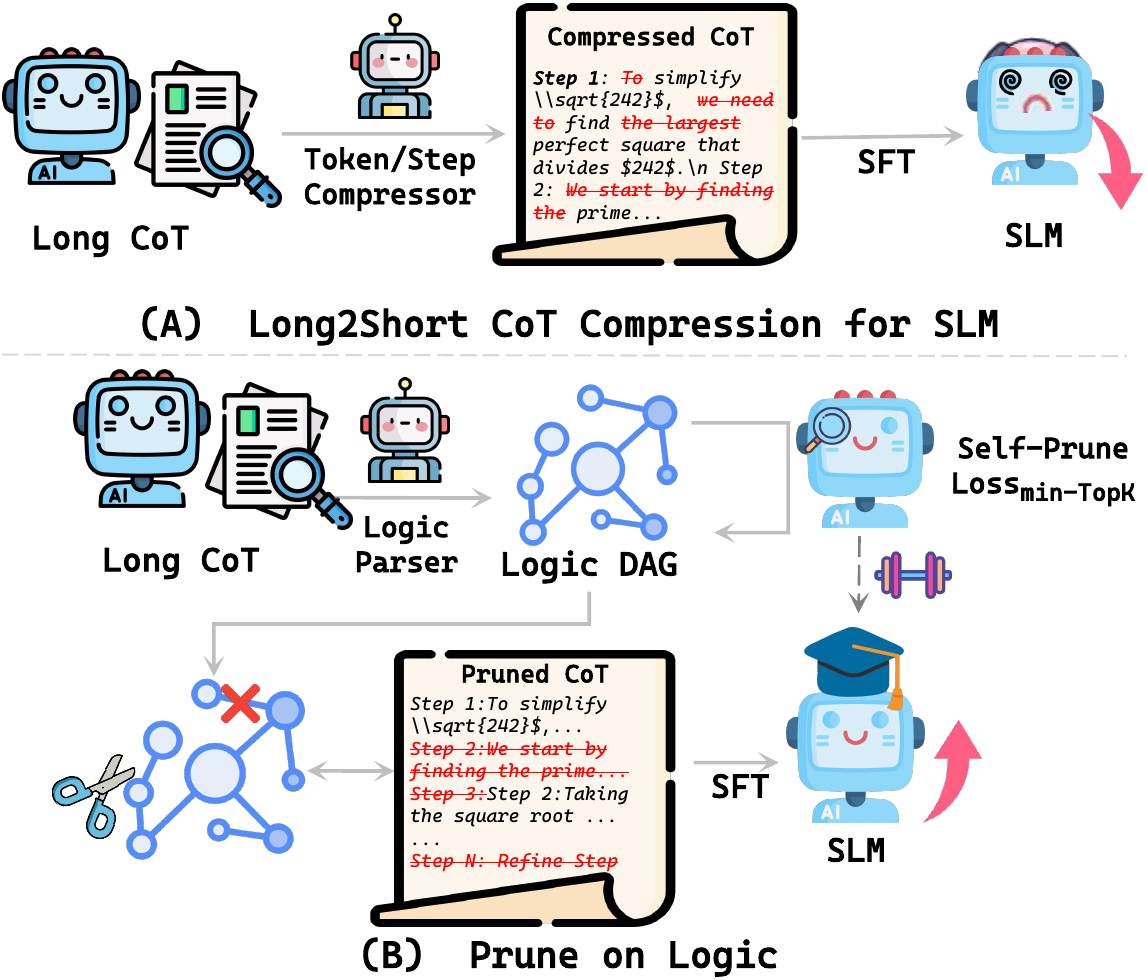}
\caption{
Comparison of token-level vs. logic-based CoT compression.  
(A) Long2Short trims surface tokens \cite{xia2025tokenskip,hou2025thinkprune}, risking semantic loss.  
(B) Prune-on-Logic preserves deductive structure via equivalence-based graph pruning.
}
\label{fig:intro}
\end{figure}

\section{Introduction}
Large Language Models (LLMs), including OpenAI-O1 \cite{jaech2024openai} and Deepseek-R1 \cite{guo2025deepseek}, have demonstrated remarkable capabilities in complex reasoning across mathematical problem-solving \cite{zhang2024learn} and logical reasoning \cite{wan2024logicasker}. These models have evolved from Short Chain-of-Thought (CoT) reasoning with concise, stepwise logic \cite{wei2022chain,yang2024buffer,wu2024beyond} to Long-CoT reasoning with deeper chains \cite{jin2024impact,jiang2024comt} via iterative exploration \cite{bi2025cot, yang-etal-2025-markov} and reflection \cite{zhang2024learn}. While Long-CoT reasoning significantly boosts reasoning capacity, it also introduces considerable computational overhead, with extended reasoning chains amplifying inference latency, memory demands driven by key-value caching, and the quadratic computational cost of attention \cite{zhang2025lightthinker}—effects that are magnified by both model scaling \cite{zhang2024scaling,guo2025deepseek} and test-time thinking scaling \cite{yang-etal-2025-markov}, severely limiting LLMs’ effectiveness in real-world complex reasoning tasks under long-thought reasoning pattern.

To address these resource constraints, recent research has increasingly focused on transferring the Long-CoT reasoning abilities of large-scale language models into Smaller Language Models (SLMs) via CoT distillation \cite{ranaldi2024aligning,bi2025prism} to perform complex long-context reasoning with greater efficiency \cite{shridhar2023distilling,ranaldi2024aligning,li2025catp}. However, a key challenge remains: SLMs face inherent limitations in acquiring long-context reasoning due to their smaller capacities. Unlike traditional distillation approaches, which primarily focus on performance, recent work emphasizes how SLMs can learn to perform complex long-context reasoning despite their capacity constraints \cite{li2025small}. Token-level compression via prompting \cite{xia2025tokenskip} or LLMLingua \cite{pan2024llmlingua} has been shown to be lossy for finetuning \cite{han2024token} or RLHF \cite{hou2025thinkprune}, frequently discarding critical information necessary for accurate long-context reasoning \cite{zhang2025lightthinker}. While existing approaches such as length-harmonizing decoding \cite{luo2025o1} and hybrid training on mixed-length data \cite{li2025small} improve the likelihood of generating long or short CoT outputs, they largely focus on output preferences rather than fundamentally enhancing SLMs' ability to learn from long-CoT reasoning. 

In this work, we revisit long-CoT compression from a capability-centric perspective and ask: \textbf{Can Pruning Improve Reasoning?} Instead of compressing at the token level, we investigate whether pruning logically redundant or minimally impactful reasoning steps can enhance SLMs' ability to perform long-context reasoning more effectively. Our goal is to distill not just shorter outputs, but more efficient and cognitively meaningful reasoning paths that align better with SLM's limited capacity. Our contributions are summarized as follows:

\begin{itemize}
\item We propose \textbf{Prune-on-Logic}, a structure-aware compression framework that transforms Long-CoT into logic graphs and enables selective pruning of low-utility reasoning steps, guided by loss-based importance scoring.

\item We conduct a systematic comparison of three pruning strategies—All-chain, Reasoning-only, and Verification-only—and reveal that pruning all verification steps consistently improves accuracy and reduces inference cost, while other strategies degrade performance.

\item We reveal that pruning logic-bearing nodes, particularly within the verification phase, enhances reasoning effectiveness in SLMs. This aligns the reasoning process with model capacity and capability, offering a robust mechanism for scalable Long-CoT distillation.

\end{itemize}
\begin{table}[ht]
\centering
\footnotesize
\resizebox{1.0\linewidth}{!}{
\begin{tabular}{lccccccccccc}
\toprule
\textbf{Method} & \textbf{Compr.} & \textbf{Lossless} \\
\midrule
Budget \cite{han2024token} & \textit{Cut Token} & \XSolidBrush  \\
TokenSkip \cite{xia2025tokenskip} & \textit{Cut Token} & \XSolidBrush  \\
ThinkPrune \cite{hou2025thinkprune} & \textit{Cut Token} & \XSolidBrush  \\
LightThinker \cite{zhang2025lightthinker} & \textit{Cut 
 Step} & \XSolidBrush  \\
O1-Pruner \cite{luo2025o1} & \textit{no Cut} & {\Large\bm{$\checkmark$}}  \\
Mix-chain \cite{li2025small} & \textit{no Cut} & {\Large\bm{$\checkmark$}}  \\
\rowcolor[RGB]{236,244,252}
\textbf{Prune-on-Logic (Ours)} & \textit{Cut Step} & {\Large\bm{$\checkmark$}}  \\
\bottomrule
\end{tabular}}
\caption{Comparison of existing Long-CoT compression approaches for SLMs. Most methods involve lossy compression and risk discarding key reasoning steps, affecting model performance on complex tasks.}
\label{tab:compression}
\end{table}

\section{Related Work}
\paragraph{LLM as a Graph Parser.}
To better manage long CoT reasoning, we decompose the original long CoT into a step-level logic graph for structured reasoning, following \cite{pandey2025adaptive}. Recent advances in structured reasoning via CoT \cite{zhang2025ratt,zhang2025sr} and self-correction \cite{zhang2025self} have improved LLM-based parsers and revealed their ability to represent logical structures as graph-based reasoning processes \cite{luo2025deconstructing}. Unlike prior methods that extract isolated sub-information, we preserve the original sentence's logical flow by decomposing it into ordered reasoning steps. Each step undergoes a verification phase to maintain fidelity to the original content. This design yields a semantically equivalent and structurally organized representation.

\begin{figure*}[ht]
  \centering
  \includegraphics[width=\linewidth]{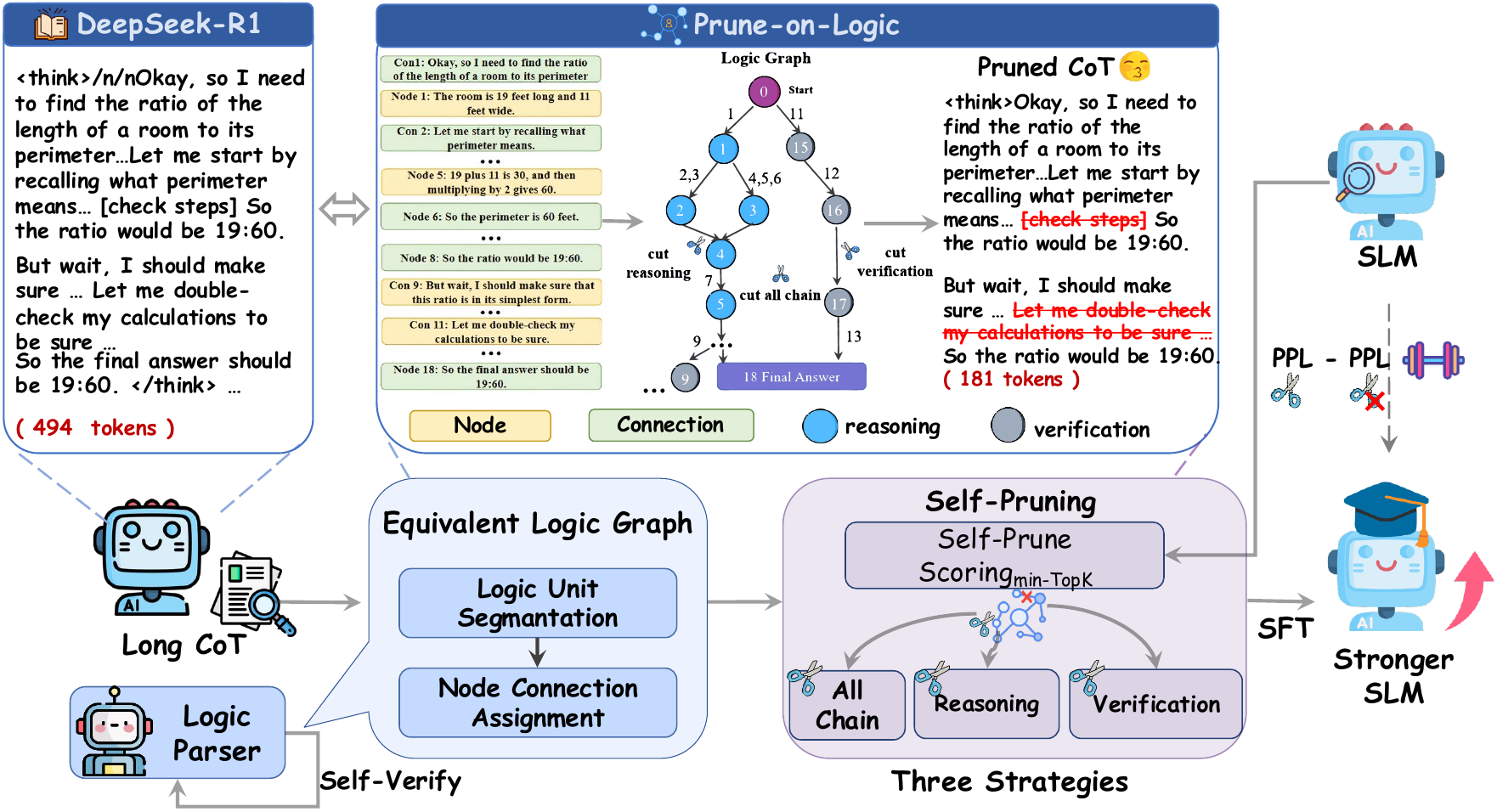}
  \caption{
Overview of the Prune-on-Logic framework. Given a Long-CoT sequence, we build a logic graph that captures deductive dependencies and, through self-pruning under logical equivalence constraints, scores and eliminates low-impact reasoning steps. We explore three pruning strategies—\textit{All-chain}, \textit{Reasoning-only}, and \textit{Verification-only}—and fine-tune SLMs on the compressed CoTs to enhance both efficiency and reasoning robustness.
  }
  \label{fig:model}
\end{figure*}

\paragraph{Long2short CoT Compression for SLMs.} Existing compression methods for SLMs are inherently lossy. Fine-tuning token-level compression via perplexity \cite{han2024token}, prompting strategies \cite{hou2025thinkprune}, step-level compression \cite{zhang2025lightthinker}, or RLHF on length-harmonizing datasets \cite{luo2025o1} often degrade reasoning capabilities. As shown in Table~\ref{tab:compression}, existing methods, while improving efficiency, risk discarding critical reasoning information \cite{hou2025thinkprune} necessary for complex tasks. Our work primarily revisits how step-level compression can enhance SLMs. As such, we do not consider comparisons with alternative approaches such as data mixing tricky strategies \cite{li2025small} or budget prompting strategies \cite{han2024token}. To the best of our knowledge, this is the first work to systematically explore perplexity-aware step-level pruning, offering new insights into how lossy pruning affects different CoT stages (e.g., reasoning, verification).

\section{Prune-on-Logic}
Inspired by the complex graph-like connectivity in biological brains \cite{friston2008hierarchical}, we adopt a structure-aware view of reasoning data following the advice of \citet{zhang2025lightthinker,luo2025deconstructing}. We propose \textbf{Prune-on-Logic} involving: (1) logic graph construction (§\ref{subsec:construct}), (2) step-level importance scoring via predictive loss and ranking-based pruning of low-impact steps (§\ref{subsec:selfprune}). We systematically explore three pruning strategies over different reasoning phases, aiming to compress Long-CoT without degrading inference quality, and revisit the question: Can pruning improve reasoning, when compression is guided not by length but by capability alignment? 
These strategies target different phases of the reasoning process: full sequences, core derivations, and verification tails. This allows us to probe not only \textit{where} pruning is most effective, but also \textit{what} should be pruned.

Specifically, we explore two granularities of pruning: (1) removing individual logic nodes, which represent discrete reasoning steps, and (2) removing edges, i.e., weakening or severing weakly informative logical connections between steps. This distinction is important: pruning a node removes its content entirely, while pruning an edge changes the dependency structure and may isolate redundant subchains. By analyzing both forms of pruning across different reasoning roles, we revisit a fundamental question:\textit{ Can logic-aware pruning compress CoT traces not by truncation, but by selectively removing parts that least support inference—across steps, stages, and rhetorical roles?}

\subsection{Equivalent Logic Graph Reconstruction}
\label{subsec:construct}
Long chain-of-thought (Long-CoT) sequences often contain rich, multi-step reasoning interleaved with rhetorical scaffolding \cite{friston2008hierarchical}. While their original form is linear, the underlying logic is inherently graph-structured \cite{luo2025deconstructing}.  
To make such structure explicit without altering the content, following \citet{luo2025deconstructing}, we map each CoT sequence as a semantically equivalent logic graph that preserves all deductive steps but encodes them as a directed acyclic graph (DAG). This transformation does not alter the meaning of the reasoning; rather, it restructures the sentence-level logic into a graph topology that reflects the actual flow of inference.

\subsection{Equivalent Logic Graph Reconstruction}
\label{subsec:construct}

Long chain-of-thought (Long-CoT) sequences encode rich, multi-step reasoning in a linear textual form. However, their underlying semantic structure is inherently graph-based \cite{luo2025deconstructing}. To reveal this structure without altering the original content, we transform each CoT into a semantically equivalent directed acyclic graph (DAG), denoted as $\mathcal{G} = (\mathcal{N}, \mathcal{E})$, where nodes represent deductive steps and edges denote logical dependencies.
\paragraph{Logic Unit Segmentation.}
Given a CoT sequence $\mathcal{S} = \{s_1, s_2, \dots, s_n\}$, we prompt a large language model (LLM) to segment it into two disjoint subsets using a classification prompt $P_{\text{logic}}$ (see Appendix~\ref{subapp:logic}):
\begin{itemize}[itemsep=0em, parsep=0pt, topsep=0pt]
    \item $\mathcal{N}$: original sentences that express reasoning or verification steps, such as symbolic operations, numerical derivations, or inference-bearing conclusions;
    \item $\mathcal{C}$: original sentences that function as rhetorical connectors, including fillers, transitions, or metacognitive phrases.
\end{itemize}

This yields a structured annotation of the original sequence:
\begin{equation}
(\mathcal{N}, \mathcal{C}) = \mathcal{LLM}(\mathcal{S}; P_{\text{logic}})
\end{equation}
Each sentence is assigned a role label, and its original position is preserved for downstream reference.

\begin{algorithm}[t]
\caption{Logic-Aware Self-Pruning}
\label{alg:logic-prune}
\begin{algorithmic}[1]
\REQUIRE CoT sequence $\mathcal{S} = \{s_1, s_2, \dots, s_n\}$, pruning ratio $\gamma \in (0, 1)$, pre-Sft LLM $\mathcal{SLM}$
\ENSURE Pruned sequence $\mathcal{S}_{{pruned}}$

\STATE \textbf{Node Segmentation:} Partition $\mathcal{S}$ into logic steps $\mathcal{N}$ and rhetorical links $\mathcal{C}$
\STATE \textbf{Graph Extraction:} Construct latent DAG $\mathcal{G} = (\mathcal{V}, \mathcal{E})$ by selecting minimal support sets $F_i$ for each $n_i \in \mathcal{N}$ (Eq.~\ref{eq:support_set}) and assigning edge for each connection $c_i \in \mathcal{C}$
\STATE \textbf{Pruning Candidate Identification:} Extract linear candidates $\mathcal{N}_{{prune}} \subseteq \mathcal{N}$ such that $|F_{n_i}| = 1$ and $|T_{n_i}| = 1$
\FORALL{$n_i \in \mathcal{N}_{{prune}}$}
    \STATE Let $to_i$ be the successor of $n_i$
    \STATE Define step span $[p_{\text{st}}, p_{\text{ed}}]$ from $n_i$ to $to_i$
    \STATE Compute semantic utility $\text{Score}_i$ using $\mathcal{SLM}$ (Eq.~\ref{eq:retain}--\ref{eq:prune})
\ENDFOR
\STATE \textbf{Score Ranking:} Sort $\mathcal{N}_{prune}$ by $\text{Score}_i$ in ascending order.
\STATE \textbf{Step Selection:} Prune top-$k = \lfloor \gamma \cdot |\mathcal{N}| \rfloor$ low-impact nodes from $\mathcal{N}_{{prune}}$
\STATE \textbf{Sequence Reconstruction:} Form $\mathcal{S}_{{pruned}}$ by removing pruned steps from $\mathcal{S}$ while preserving original step order
\RETURN $\mathcal{S}_{{pruned}}$
\end{algorithmic}
\end{algorithm}

\paragraph{Node Connection Assignment.}
Based on the segmented units, we construct a logic graph $\mathcal{G} = (\mathcal{V}, \mathcal{E})$, where $\mathcal{V} = \mathcal{N}$ and $\mathcal{C} \subseteq \mathcal{E}$. For each reasoning node $n_i \in \mathcal{N}$, we prompt the LLM (via $P_{{conn}}$ detailed in Appendix~\ref{subapp:con}) to select a minimal set of prior nodes $F_i \subseteq \mathcal{N}_{<i}$ that justify its derivation. This induces a directed edge set:
\begin{equation}
\mathcal{E} = \bigcup_{n_i \in \mathcal{N}} \{(f, n_i) \mid f \in F_i\}
\end{equation}
where support sets are predicted via:
\begin{equation}
\label{eq:support_set}
F_i = \mathcal{LLM}(n_i; \mathcal{N}_{<i}, P_{{conn}})
\end{equation}
Subsequently, we map each connection to a graph edge, assuming its target node is the immediate successor in the text sequence. To improve connection accuracy, we incorporate a self-verification step in which the LLM is re-prompted to reassess or validate each predicted $F_i$. This iterative check enhances robustness against spurious links and promotes logical coherence detailed in Appendix~\ref{subapp:con}. The resulting graph $\mathcal{G} = (\mathcal{V}, \mathcal{E})$ not only reconstructs the latent deductive structure of the original CoT sequence but also records explicit mappings between graph nodes and their corresponding sentence indices. This enables symbolic reasoning and text-level alignment to be jointly exploited in downstream tasks. We adopt the semantic equivalence hypothesis \cite{luo2025deconstructing}—that $\mathcal{G}$ faithfully preserves the full deductive semantics of $\mathcal{S}$—as further detailed in Appendix~\ref{app:semantic_equivalence}.

\subsection{Self-Pruning via Loss-Based Ranking}
\label{subsec:selfprune}

Building on the constructed logic graph $\mathcal{G} = (\mathcal{V}, \mathcal{E})$, we further introduce a loss-based ranking mechanism \cite{pan2024llmlingua} that estimates each node's semantic contribution via token-level perplexity difference. Importantly, we restrict pruning to nodes that (1) have exactly one predecessor and successor in $\mathcal{G}$; (2) form a linear inference segment.

Let $n_i$ be a candidate node and $to_i$ its immediate successor in $\mathcal{G}$. We define the span $[p_{\text{st}}, p_{\text{ed}}]$ as the sentence range covering $n_i$ through $to_i$. To assess the semantic contribution of $n_i$, we compute token-level perplexity over this span using a frozen pre-SFT small language model $\mathcal{SLM}$:

\noindent\textit{\underline{Perplexity with node retained:}}
\begin{flalign}
\label{eq:retain}
\scriptsize
& \mathcal{PPL}_{\textit{retain}} = \exp\Bigg( \frac{1}{L_i} \sum_{j=p_{\text{st}}}^{p_{\text{ed}}} \sum_{k=1}^{t_j} \nonumber \\
& \quad - \log P(tok_{jk} \mid s_{<j}, \{tok_{jl}\}_{l<k}; \mathcal{SLM}) \Bigg) &
\end{flalign}
\noindent\textit{\underline{Perplexity with node pruned:}}
\begin{flalign}
\label{eq:prune}
\scriptsize
& \mathcal{PPL}_{\textit{prune}} = \exp\Bigg( \frac{1}{L_i} \sum_{j=p_{\text{st}}}^{p_{\text{ed}}} \sum_{k=1}^{t_j} \nonumber \\
& \quad - \log P(tok_{jk} \mid s_{<j} \setminus n_i, \{tok_{jl}\}_{l<k}; \mathcal{SLM}) \Bigg) &
\end{flalign}
Then, the self-pruning semantic utility score is:
\begin{equation}
    \textit{Score}_i = \mathcal{PPL}_{{prune}} - \mathcal{PPL}_{{retain}}
\end{equation}
Here, $\textit{Score}_i$ reflects how much the removal of $n_i$ degrades the model’s ability to predict downstream tokens. Lower scores indicate that $n_i$ contributes little to semantic continuity and may be safely pruned. We sort all eligible nodes by $\textit{Score}_i$ and remove the top-$k$ lowest, where $k = \lfloor \gamma \cdot |\mathcal{N}| \rfloor$ and $\gamma$ is a pruning ratio hyperparameter and get the final pruned CoT $S_{pruned}$ (See Appendix~\ref{subapp:data} for a Prune-on-Logic data example.)

\paragraph{Pruning Strategies.}
To explore how pruning affects reasoning, we define three logic-aware strategies based on functional roles of nodes in the CoT process. The \textbf{All-chain} strategy removes low-scoring nodes from the entire logic graph, balancing breadth and depth of reasoning. This acts as a general compression baseline. The \textbf{Reasoning-only} strategy prunes derivational steps—such as intermediate calculations—while preserving final conclusions and verification. This reflects a conservative approach that favors robustness over maximal compression. Conversely, the \textbf{Verification-only} strategy targets self-checking or reflective utterances (e.g., “Let’s verify that”), aiming to reduce rhetorical redundancy while leaving the computational core intact. Each strategy produces a pruned sequence $\mathcal{S}_{{pruned}}$ by deleting selected nodes from the original CoT $\mathcal{S}$ in order. The number of deletions is controlled by a pruning ratio $\gamma \in (0, 1)$, which determines the top-$k$ nodes to remove based on their semantic utility scores (§\ref{subsec:selfprune}). This explicit control allows us to examine not only \emph{where} pruning is most effective, but also \emph{how much} pruning can be tolerated before degrading reasoning quality.

\paragraph{Pruning Component.}
Beyond determining \textit{where} to prune within the reasoning process, we also consider \textit{what} to prune. Specifically, we explore two levels of granularity: \textbf{node-level} and \textbf{connection-level} pruning. These correspond to distinct components of a CoT sequence.
\begin{itemize}[itemsep=0em, parsep=0pt, topsep=0pt]
    \item \textbf{Node-level pruning} removes entire reasoning steps $n_i \in \mathcal{N}$, such as symbolic derivations and their corresponding connections.
    \item \textbf{Connection-level pruning} suppresses rhetorical units $c_i \in \mathcal{C}$, such as discourse fillers and stylistic transitions.
\end{itemize}

Both types are evaluated under the same loss-based scoring mechanism introduced in §\ref{subsec:selfprune}. For each candidate unit, we compute its semantic utility $\textit{Score}_i$ using the differential loss with and without the unit. This unified formulation enables selective removal of both deductive and rhetorical components. Rhetorical connections are retained only when their removal would harm the model’s predictive confidence. Together, the two pruning axes—\textit{strategy scope} (§\ref{subsec:selfprune}) and \textit{component granularity}—form a unified, structure-aware framework. This design allows us to ask: \emph{What parts of a reasoning chain can be removed without impairing performance, and how does this vary under different resource constraints?}
\begin{table*}[t!]
\centering
\resizebox{1.0\linewidth}{!}{
\begin{tabular}{lccccccccccc}
\toprule
\multirow{2}{*}{\makecell{\textbf{Pruning Ratio}\\ ~$\boldsymbol{\gamma}$}}
& 
\multicolumn{3}{c}{\textbf{In-Domain}} & 
\multicolumn{2}{c}{\textbf{Out-of-Domain}} & 
\multicolumn{2}{c}{\textbf{Overall Performance}} \\
\cmidrule(lr){2-4} \cmidrule(lr){5-6} \cmidrule(lr){7-8}
& AMC23$\uparrow$ & AIME$\uparrow$ & MATH500$\uparrow$ & GSM8K$\uparrow$ & BBH$\uparrow$ & Avg. Acc.$\uparrow$ & Avg. Token ($\Delta$)$\downarrow$ \\
\midrule 
\rowcolor{gray!8}
\multicolumn{8}{c}{DeepSeek-R1-Distill-Llama-8B~\citep{guo2025deepseek}} \\
\midrule
Original & 65.0 & 25.0 & 81.2 & 80.4 & 37.0 & \underline{57.7} & 4555.9 \\
\midrule

\multicolumn{8}{l}{\textit{\textbf{Pruning Nodes~over All Chains}}} \\
\;\; Cut 0.5 &   40.0    & 25.5 & 77.4 &   79.4    &   37.9    & 52.0  & {4707.1 ( +3.32\%)} \\
\;\; Cut 0.7 &   37.5    & 30.0 & 72.6 &   71.1    &   25.0    & 47.2 & {4824.7 (+5.90\%)} \\
\;\; Cut 0.9 &   40.0    & 28.4 & 78.4    &   78.4    & 34.3 & 51.9 & {4846.2 (+6.37\%)} \\
\;\; Cut All & 5.0  &   3.3   &  14.4  &   13.4    &    15.7   &  10.4 & {475.4 (-89.6\%)}  \\

\multicolumn{8}{l}{\textit{\textbf{Pruning Nodes~over Reasoning Chains}}} \\
\;\; Cut 0.5 &    50.0   & 26.7 & 81.0 &    76.3   &    37.9   & 54.4 & {4755.0 (+4.37\%)} \\
\;\; Cut 0.7 &   42.5    & \textbf{33.4} & 80.8 &   \underline{82.5}    &     29.6  & 53.8  & {4779.9 (+4.92\%)} \\
\;\; Cut 0.9 &   47.5    &  \underline{30.0} & 76.6 &      80.4 &   33.3    & 53.6 & {4707.6 (+3.33\%)} \\
\;\; Cut All & 35.0 &  5.0 & 51.4 &   73.2    &  25.9     & 38.1 & {1681.5 (-63.1\%)}  \\

\multicolumn{8}{l}{\textit{\textbf{Pruning Nodes~over Verification}}} \\
\;\; Cut 0.5 & 57.5  & 26.7 & \underline{83.2} &    80.0   &   35.2    & 56.5 & {4786.2 (+5.05\%)} \\
\;\; Cut 0.7 & 60.0  & 28.3 & 82.8 &    79.0   &   35.2    & 57.1 & {4392.6 (-3.58\%)} \\
\;\; Cut 0.9 &   37.5    & \underline{30.0} & 82.2 &     81.4  &    37.1   & 53.6 & {4428.3 (-2.80\%)} \\
\rowcolor[RGB]{236,244,252}
\;\; Cut All & \textbf{80.0}  & \underline{30.0} & \textbf{84.2} & \textbf{82.5} &   \textbf{38.0}   & \textbf{62.9} & {4295.2 (-5.72\%)} \\
\midrule

\rowcolor{gray!8}
\multicolumn{8}{c}{DeepSeek-R1-Distill-Qwen-7B~\citep{guo2025deepseek}} \\
\midrule
Original & 60.0 & 33.3 & 82.8 & 88.6 & 20.4 & 57.0 & 4344.3 \\
\midrule

\multicolumn{8}{l}{\textit{\textbf{Pruning Nodes~over All Chains}}} \\
\;\; Cut 0.5 &   47.5    & 33.4 &   85.8    &     88.7  &   12.9    & 53.7 & {4183.9 (-3.69\%)} \\
\;\; Cut 0.7 &   57.5     & 38.3 & 85.4 &  85.6     &   19.4    & 57.2 & {4249.0 (-2.19\%)} \\
\;\; Cut 0.9 &   55.0    & \underline{41.7} & 83.2 &    \textbf{89.7}   &    17.6   & 57.4  & {4306.8 (-0.86\%)} \\
\;\; Cut All &  15.0 & 3.3 & 14.8 &   36.1    &  21.3     & 18.1  &  {462.0 (-89.4\%)} \\

\multicolumn{8}{l}{\textit{\textbf{Pruning Nodes~over Reasoning Chains}}} \\
\;\; Cut 0.5 &    47.5   & 33.4 &    85.4  &   86.6    &   \underline{25.0}    & 55.6  &  {4066.7 (-6.39\%)} \\
\;\; Cut 0.7 &   60.0    & \underline{41.7} & \textbf{87.6} &   87.4    &   18.5   &  \underline{59.0} &  {4210.6 (-3.08\%)} \\
\;\; Cut 0.9 &    47.5   & \underline{41.7} & 85.0 &   87.6    &    22.2   & 56.8  & {4236.2 (-2.49\%)}  \\
\;\; Cut All & 50.0 &  3.3  & 48.8 &   81.4    &      24.0 & 41.5  &  {776.6 (-82.1\%)} \\

\multicolumn{8}{l}{\textit{\textbf{Pruning Nodes~over Verification}}} \\
\;\; Cut 0.5 &   57.5    & 33.4 &   83.8  &     \textbf{89.7}  &  22.2 & 57.3 & {3998.7 (-7.96\%)} \\
\;\; Cut 0.7 &   55.0    & \underline{41.7} & 83.0 &     \textbf{89.7}  &  21.3    &  58.1 & {4169.0 (-4.03\%)}  \\
\;\; Cut 0.9 & 65.0  & 
36.7 & \underline{86.0} &    88.7   &    18.5   &  \underline{59.0} &  {4258.5 (-1.97\%)} \\
\rowcolor[RGB]{236,244,252}
\;\; Cut All & \textbf{72.5} & \textbf{43.3} & \underline{86.0} & \underline{89.0} &  \textbf{25.9} & \textbf{63.0} & {3931.2 (-9.51\%)}  \\
\bottomrule
\end{tabular}}
\caption{
Evaluation results across pruning strategies, pruning ratios ($\gamma$), and two distilled SLMs. We report Avg. Acc. (pass@1 accuracy averaged across benchmarks) and Comp. (relative token reduction during inference) under three logic-based pruning strategies—\textit{All-chain}, \textit{Reasoning-only}, and \textit{Verification-only}. The uncompressed LongCoT-SFT baseline ($\gamma=0.0$) serves as the reference point. Verification pruning yields the best accuracy-efficiency trade-off, while aggressive all-chain pruning degrades performance despite high compression. Appendix~\ref{app:additional} details extensive experiments on scaling and Long-CoT effects.
}
\label{tab:results}
\end{table*}

\section{Experiment}

\subsection{Experimental Setup}
\paragraph{Models and Datasets.}
To investigate whether pruning Long CoT enhances SLM reasoning, we experiment on DeepSeek-R1-Distill-Llama-8B and DeepSeek-R1-Distill-Qwen-7B~\cite{guo2025deepseek}, comparing three pruning strategies applied at different reasoning stages (all-chains, reasoning chains, and verification pruning) under pruning ratios \{0.5, 0.7, 0.9, 1.0\}, with the original supervised fine-tuning (SFT) model as the baseline. Following \citet{zhang2025lightthinker}, we perform LoRA \cite{hu2022lora} on Bespoke-Stratos-17k dataset \cite{bespoke_stratos} for training and evaluate models on in-domain test sets (AMC 2023~\cite{li2024numinamath}, AIME~\cite{li2024numinamath}, MATH500~\cite{lightman2023let}) and out-of-domain test sets (GSM8K~\cite{cobbe2021training}, Big-Bench Hard~\cite{suzgun2023challenging}). Additionally, we expand verification pruning to (1) CommonsenseQA~\cite{talmor2019CommonsenseQA} for transfer to non-math reasoning, (2) Llama3.1-8B-Instruct~\cite{grattafiori2024Llama} to assess models without Long-CoT pretraining, and (3) DeepSeek-R1-Distill-Qwen-1.5B~\cite{guo2025deepseek} for robustness under constrained capacity. Detailed results on out-of-math tasks, Long-CoT, and scaling are in Section~\ref{subsec:main_result} and Appendix~\ref{app:additional}.  

\paragraph{Implementation Details.}
To construct accurate initial logic-step datasets, we employed a three-phase directed acyclic graph (DAG) construction pipeline: node-connection differentiation, structure initialization with nodes as anchors, and directed edge assignment based on semantics (Appendix~\ref{app:prompt}). The pipeline was executed using the outstanding \text{gemini-2.5-pro-preview-05-06} model. For logic-level importance scoring, we used the pre-SFT model to compute token-level loss values (detailed in §~\ref{subsec:selfprune}). After pruning on logic-level datasets, we applied LoRA \cite{hu2022lora} on MS-Swift \cite{zhao2025swift} with default hyperparameter, a learning rate of 1e-4 for 1 epoch on one NVIDIA-A100 GPU. All results are averaged over three runs with inference temperature set to 0.0 for reproducibility.

\begin{table*}[t!]
\centering
\small
\renewcommand{\arraystretch}{1}
\resizebox{1.0\linewidth}{!}{
\begin{tabular}{lccccccccccc}
\toprule
\textbf{Pruned Component} & \textbf{AMC23}$\uparrow$ & \textbf{AIME}$\uparrow$ & \textbf{MATH500}$\uparrow$ & \textbf{GSM8K}$\uparrow$ & \textbf{BBH}$\uparrow$ & \textbf{Avg. Acc.}$\uparrow$ & \textbf{Avg. Token ($\Delta$)}$\downarrow$ \\
\midrule
\rowcolor{gray!8}
\multicolumn{8}{c}{DeepSeek-R1-Distill-Llama-8B~\citep{guo2025deepseek}} \\
\midrule
\multicolumn{8}{l}{\textit{Pruning over Verification}} \\
\;\; Cut All Connections  &  2.5 & 28.3  & 81.3  & 32.0  & 14.8    & 31.8 & {4362.9 (-4.24\%)} \\
\rowcolor[RGB]{236,244,252}
\;\; Cut All Nodes & \textbf{80.0}  & \textbf{30.0} & \textbf{84.2} & \textbf{82.5} &   \textbf{38.0}   & \textbf{62.9} & \textbf{4295.2 (-5.72\%)} \\
\midrule
\rowcolor{gray!8}
\multicolumn{8}{c}{DeepSeek-R1-Distill-Qwen-7B~\citep{guo2025deepseek}} \\
\midrule
\multicolumn{8}{l}{\textit{Pruning over Verification}} \\
\;\; Cut All Connections  &  2.5 & 28.3  & 77.2 & 30.9 & 13.0    & 30.4  & {4301.0 (-1.00\%)} \\
\rowcolor[RGB]{236,244,252}
\;\; Cut All Nodes & \textbf{72.5} & \textbf{43.3} & \textbf{86.0} & \textbf{89.0}    & \textbf{25.9} & \textbf{63.0} & \textbf{3931.2 (-9.51\%)} \\
\bottomrule
\end{tabular}}
\caption{
Comparison of pruning logical nodes vs.\ rhetorical connections within verification spans. Pruning logic-bearing nodes consistently improves reasoning accuracy and token efficiency over connection-level pruning across models and domains, underscoring the structural value of even shallow verification steps.
}
\label{tab:ablation-verification-domain}
\end{table*}

\begin{figure}[ht]
\centering
\includegraphics[width=\linewidth]{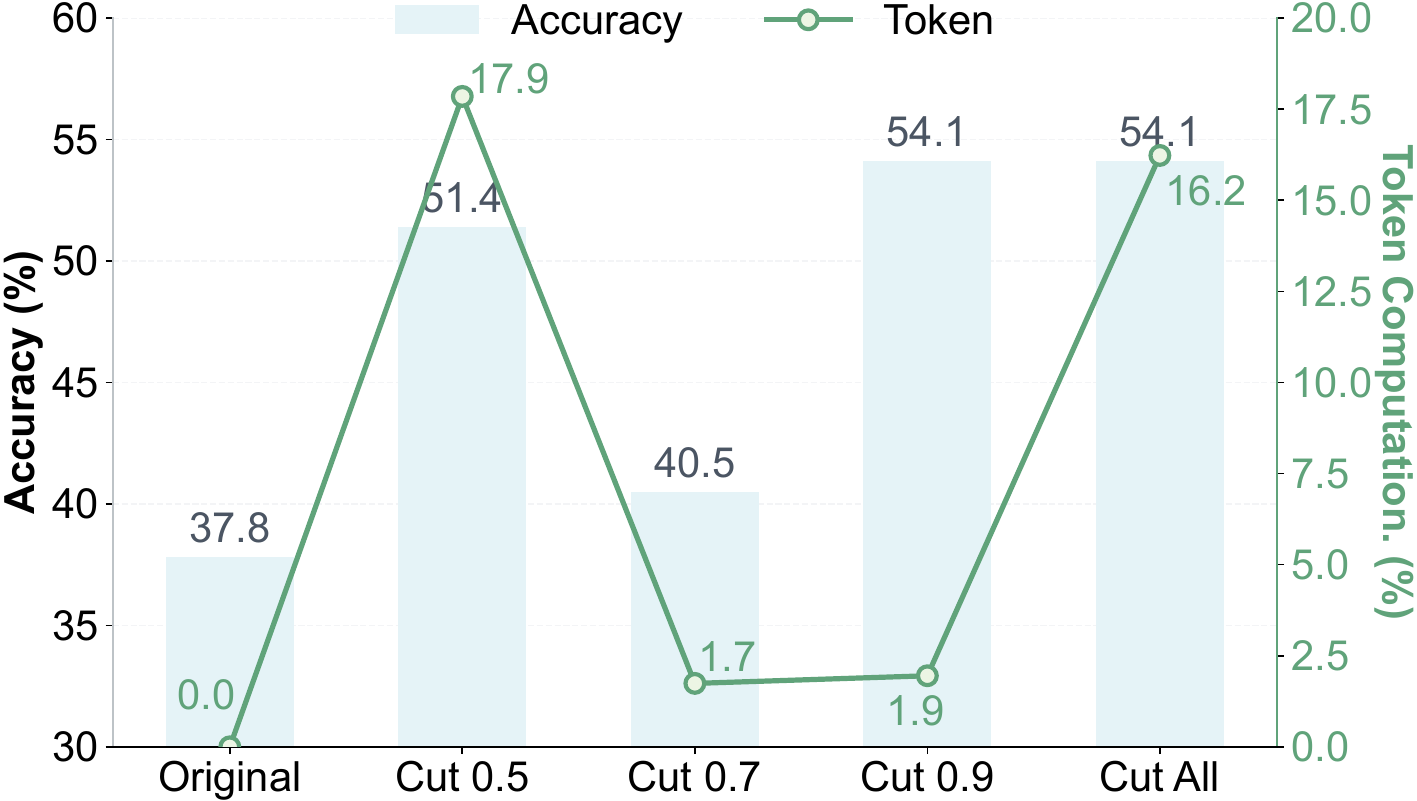}
\caption{
CommonsenseQA: Verification Pruning on R1-Distill-Qwen-7B.
}
\label{fig:commonsenseqa}
\end{figure}

\paragraph{Evaluation Metrics.}
Following \citet{zhang2025lightthinker,ye2025limo}, we evaluate performance using pass@1 accuracy ($\text{Acc.}$) under a zero-shot setting, and report average accuracy (Avg. Acc.) across all benchmarks. For efficiency, we report the average number of inference tokens (\textbf{Token}) alongside its relative change ($\Delta$) compared to the uncompressed Long-CoT SFT baseline. $\Delta$ is computed as the percentage reduction or increase in token count: {\texttt{-}} for compression (fewer tokens) and {\texttt{+}} for expansion (more tokens). This joint display captures both absolute token usage and efficiency trends under various pruning strategies.

\subsection{Main Results}
\label{subsec:main_result}
\paragraph{Can Pruning Improve Reasoning? Yes—If Structured and Aligned with Model Capacity.} To assess whether pruning from Long-CoT can improve reasoning instead of lossy learning for SLMs~\cite{han2024token,xia2025tokenskip,hou2025thinkprune}, we evaluate three pruning strategies that target all-chain, reasoning-only, and verification-only components under varying pruning ratios and model capacities. Our results reveal that pruning is not merely a compression tool, but a mechanism for aligning supervision with SLMs' capability. The key lies in what gets pruned—not how much.

\paragraph{Finding 1: Longer Isn’t Always Better: Redundancy Hurts SLMs Learning from Long-CoT.}
Despite the common belief \cite{jin2024impact} that longer CoT enhances reasoning, our results show that excessive length—especially during verification, can overwhelm small models with shallow content that adds little deductive value. As shown in Table~\ref{tab:results}, pruning verification steps ($\gamma$ = 1.0) consistently improves accuracy: DeepSeek-R1-Distill-Qwen-7B increases from 57.0\% to 63.0\% (+6.0), and DeepSeek-R1-Distill-Llama-8B rises from 57.7\% to 62.9\% (+5.2\%) while reducing tokens by 9.51\% and 5.72\%, respectively. These results suggest that SLMs are less constrained by the availability of long reasoning traces than by their capacity to filter irrelevant or repetitive steps aligned with the premise of \citet{luo2025deconstructing}. Verification, often reiterating previous answers, is the most dispensable during Long-CoT distillation. Pruning verification steps helps the model focus on essential reasoning while reducing computational load without sacrificing logical coverage.

\paragraph{Finding 2: What We Prune Matters More Than How Much.}
Pruning is not simply a matter of reducing length; it's about preserving semantic function. Our empirical analysis across three pruning strategies (Table~\ref{tab:results}) shows that indiscriminate removal of CoT steps severely degrades performance: DeepSeek-R1-Distill-Qwen-7B drops from 57.0\% to 18.1\% (–38.9), and DeepSeek-R1-Distill-Llama-8B from 57.7\% to 10.4\% (–47.3). In contrast, selectively pruning verification steps improves both accuracy and efficiency (+6.0 for Qwen-7B, +5.2 for Llama-8B), while moderate pruning of reasoning chains ($\gamma$=0.5) still reduces performance (–3.3 and –1.4). These results indicate that small models rely on reasoning traces to acquire inductive patterns beyond final answers. Pruning must respect this structural dependency: verification steps are often reiterative and safe to remove, but reasoning steps provide the logical backbone for learning. Supervision must prioritize semantic salience over superficial verbosity.

\begin{figure}[t]
\centering
\includegraphics[width=\linewidth]{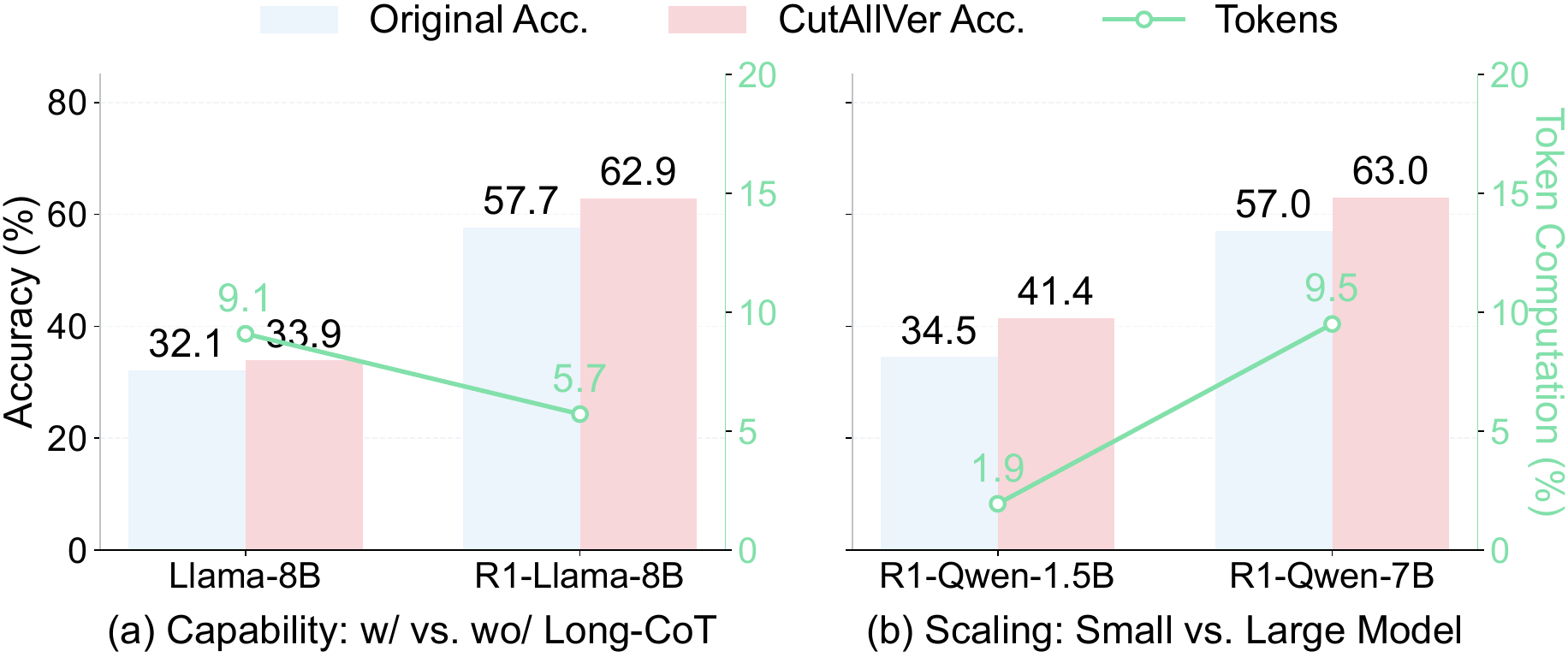}
\caption{
All-Verification pruning enhances accuracy while reducing token usage. (a): Models with (R1-Distill-Llama-8B) and without (Llama3.1) Long-CoT capability both improve, showing generality across reasoning supervision. (b): Larger R1-Distilled-Qwen models (7B) gain more from pruning than smaller ones (1.5B), highlighting benefits scale with capacity.}
  \label{fig:analysis}
\end{figure}

\paragraph{Finding 3: Verification Steps Are Structurally Dispensable—If Nodes, Not Connections, Are Removed.}
To explore finer-grained pruning targets, we compare the removal of logic-bearing \emph{nodes} vs.\ rhetorical \emph{connections} in the verification span (Table~\ref{tab:ablation-verification-domain}). Strikingly, removing verification nodes yields strong accuracy (62.9\% and 63.0\% on DeepSeek-R1-Distill-Llama-8B and DeepSeek-R1-Distill-Qwen-7B), while pruning only rhetorical connections drops accuracy below 32\%. This highlights that even shallow logic steps provide essential semantic support for predictions, while connectors offer minimal utility. Thus, pruning must be semantically aligned, not merely syntactic.

\paragraph{Finding 4: All-Verification Pruning Improves Reasoning Across Domains, Architectures, and Capacities.} 
We evaluate the robustness of verification pruning on (i) out-of-math tasks (CommonsenseQA \cite{talmor2019CommonsenseQA}), (ii) models without Long-CoT pretraining (Llama3.1-8b-Instruct \cite{grattafiori2024Llama}), and (iii) smaller-capacity SLMs (Deepseek-R1-Distill-Qwen-1.5B-Instruct \cite{guo2025deepseek}). \textbf{(1) All-verification also works on out-of-math tasks.} As shown in Figure~\ref{fig:commonsenseqa}, pruning boosts the 7B model's accuracy on CommonsenseQA from 37.8\% to 54.1\%, with 16.2\% fewer tokens, outperforming all other pruning strategies. 
Besides, \textbf{(2) Verification pruning improves reasoning across both CoT-trained and non-CoT models, though token efficiency varies with prior supervision.}  
Figure~\ref{fig:analysis} (a) shows that both Llama3.1 without Long-CoT and R1-Distilled-Llama with Long-CoT benefit from pruning, with accuracy gains of +5.2 and +1.8, respectively. However, token compression differs: Llama3.1 sees a 9.1\% reduction, while R1-Llama achieves 5.7\%. This suggests that models without Long-CoT generate more extraneous reasoning paths, making pruning more aggressive. In contrast, CoT-supervised models already optimize reasoning, leaving less redundant content. 
Moreover, \textbf{(3) Pruning shows a scaling effect, improving more with larger models.} On R1-Qwen models (Figure~\ref{fig:analysis} (b)), the 7B variant achieves similar accuracy gains (+6.0 vs. +6.9) but higher token compression (9.5\% vs. 1.9\%) than the 1.5B model. Larger models produce more redundant reasoning, which verification pruning can compress more effectively while maintaining correctness. Overall, across models, datasets, and pruning ratios, all verification-only pruning consistently yields the best trade-off between accuracy and efficiency. In contrast, reasoning pruning risks over-compression, and all-chain pruning collapses logic altogether. These results suggest that structured, logic-aware pruning is not just a token-reduction strategy, but a way to tailor supervision to match the abstraction capacity of SLMs, enhancing generalization and interpretability.
\section{Conclusion}
We revisit Long-CoT compression through a capability-aligned perspective and introduce Prune-on-Logic, a framework that transforms reasoning traces into logic graphs and selectively prunes low-utility reasoning steps. Unlike traditional length-based compression, our approach improves reasoning by removing redundant components, especially in the verification phase, enhancing both accuracy and efficiency. Extensive empirical experiments across three logic-aware pruning strategies demonstrate that pruning verification steps consistently boosts model performance, with larger models benefiting more due to redundant reasoning structures. Our findings highlight that logic-based pruning is not only an efficient compression technique but also an effective tool for strengthening small language models, paving the way for scalable and high-quality Long-CoT reasoning distillation.
\section*{Limitations}
While our work provides the first structured comparison of three logic-aware pruning strategies—All-chain, Reasoning-only, and Verification-only—under a capability-aligned lens, it assumes access to high-quality logic graphs and verification labels, which may limit applicability in fully unsupervised or noisy domains. Our pruning operates at the step level without explicitly modeling fine-grained symbolic dependencies or temporal coherence beyond local spans. Moreover, while we focus on small language models (SLMs), how pruning interacts with medium- or large-scale models remains underexplored. Finally, although we frame pruning as a form of lossless structural compression, some utility loss may still occur in highly entangled reasoning chains where step interdependencies are subtle and hard to disentangle.

We view this as an initial step toward principled CoT simplification, and future work may extend our framework to dynamic pruning at inference time \cite{wang2025every}, graph-based consistency enforcement, and hybrid supervision regimes.



\bibliography{custom}
\newpage
\appendix

\setcounter{tocdepth}{-1}
\addtocontents{toc}{\protect\setcounter{tocdepth}{2}}

\section*{Appendix of Prune-on-Logic}

This supplementary material provides in-depth insights into our Prune-on-Logic method, covering additional details, results and analysis. The appendix is organized as follows:

\tableofcontents

\newpage

\section{Prompt Details for Equivalent Logic Graph}
\label{app:prompt}
This section presents the prompt designs for constructing equivalent logic graphs from Long-CoT reasoning. It includes two core components: logic unit segmentation (\ref{subapp:logic}) and node connection assignment (\ref{subapp:con})

\subsection{Logic Unit Segmentation Prompt ($P_{logic}$)}
\label{subapp:logic}

\begin{center}
\begin{minipage}{\textwidth}
\centering
\begin{tcolorbox}[width=\textwidth, colback=white, colframe=Black, title=Logic Unit Segmentation Prompt ($P_{logic}$)]
\textbf{\#\#\# Instruction \#\#\#} \\
You are an expert in mathematical reasoning and logical discourse parsing. Given a problem-solving passage involving computation, your task is to segment the passage into discrete reasoning steps and commentary, then classify each segment as either a \textbf{node} or a \textbf{connection}, following a structured logic analysis framework.

\textbf{\#\#\# Node Identification \#\#\#}
\begin{itemize}[itemsep=0em, parsep=0pt, topsep=0pt]
    \item Mark segments that include calculations, algebraic manipulations, definitions, or logical conclusions.
    \item Always mark the final boxed answer as a node.
    \item Examples: symbolic equations, variable definitions, result statements.
\end{itemize}

\textbf{\#\#\#  Connection Detection \#\#\# }
\begin{itemize}[itemsep=0em, parsep=0pt, topsep=0pt]
    \item Identify segments that contain narrative transitions, setup, intent, commentary, or rhetorical expressions.
    \item These do not affect the logic or math derivation.
    \item Examples: “Let’s simplify this…”, “Now I need to check…”, “So far, we’ve shown…”
\end{itemize}

\textbf{\#\#\# Segment Splitting and Granularity \#\#\#}
\begin{itemize}[itemsep=0em, parsep=0pt, topsep=0pt]
    \item If a sentence mixes logic and commentary, split it into parts.
    \item Use punctuation (commas, periods) as natural split points.
    \item Label each part accordingly.
\end{itemize}

\textbf{\#\#\# Output Format \#\#\# }
\begin{verbatim}
<node begin>Logical computation or reasoning here</node end>
<connection begin>Narrative or 
commentary here</connection end>
\end{verbatim}
Preserve original sentence order. Strictly follow the format—do not add explanations. \\
\textbf{Case Examples:} \\
\textbf{Example 1:} \\
Input: \{LongCoT-1\}
Output: \{Logic Unit Segmentation-1\} \\
\textbf{Example 2:} \\
Input: \{LongCoT-2\}
Output: \{Logic Unit Segmentation-2\}
\end{tcolorbox}
\begin{center}
\captionof{figure}{Logic Unit Segmentation Prompt. ($P_{logic}$)}
\end{center}
\label{prompt_logical}
\end{minipage}
\end{center}

\clearpage
\subsection{Node Connection Assignment Prompt ($P_{{conn}}$)}
\label{subapp:con}
\begin{center}
\begin{minipage}{\textwidth}
\centering
\begin{tcolorbox}[width=\textwidth, colback=white, colframe=Black, title=Node Connection Assignment Prompt ($P_{{conn}}$)]
\textbf{\#\#\# Instruction \#\#\#} \\
You are an expert in logical graph construction and coherence analysis. Given a list of logic units that have been segmented and labeled as \texttt{<node>} or \texttt{<connection>}, your task is to assign directed edges between nodes based on logical dependence, using supporting \texttt{<connection>} units as justification.

\textbf{\#\#\# Edge Assignment Rules \#\#\#}
\begin{itemize}[itemsep=0em, parsep=0pt, topsep=0pt]
    \item Draw a directed edge from node A to node B if A provides information, computation, or reasoning that B builds on.
    \item Use nearby connection segments to justify the link, ensure that the target node is the immediate successor of the selected connection in the text.
    \item Do not connect unrelated nodes; avoid spurious links.
\end{itemize}

\textbf{\#\#\# Self-Verification Phase \#\#\#}
\begin{itemize}[itemsep=0em, parsep=0pt, topsep=0pt]
    \item After assigning all edges, perform a self-verification step:
    \begin{itemize}
        \item Check if every non-initial node has at least one valid predecessor.
        \item Ensure there are no circular dependencies unless logically necessary.
        \item Flag any node with unclear inputs or outputs.
    \end{itemize}
\end{itemize}

\textbf{\#\#\# Output Format \#\#\#}
\begin{verbatim}
<edge from=N1 to=N2>Supported by:  
"transition text or rationale"</edge>
<edge from=N2 to=N3>Supported by: 
"implicit logic or structure"</edge>

<verify>
Node N2: OK
Node N3: Missing link justification
Cycle detected: N4 → N5 → N4
</verify>
\end{verbatim}

Be precise in edge justification. Use natural language from the original text when available. Do not invent content beyond the given segments. Self-verification must be strict. \\

\textbf{Case Examples:} \\
\textbf{Example 1:} \\
Input: \{LongCoT-1 + Logic-Segments-1\}
Output: \{Edge-1\} \\
\textbf{Example 2:} \\
Input: \{LongCoT-2 + Logic-Segments-2\}
Output: \{Edge-2\} \\
\end{tcolorbox}
\begin{center}
\captionof{figure}{Node Connection Assignment Prompt ($P_{{conn}}$).}
\end{center}
\label{prompt_connection}
\end{minipage}
\end{center}

\clearpage

\section{Semantic Equivalence}
\label{app:semantic_equivalence}

Under the autoregressive factorization of language models, the likelihood of a Long-CoT sequence 
$\mathcal{S} = \{s_1, \dots, s_T\}$ can be expressed as:
\begin{equation}
P(\mathcal{S}) = \prod_{t=1}^T P(s_t \mid s_{<t}),
\end{equation}


We define a deterministic transformation $\phi: \mathcal{S} \to \mathcal{G}$, where $\mathcal{G} = (\mathcal{V}, \mathcal{E})$ is a directed acyclic graph (DAG). $\phi$ enjoys the strong LLM’s instruction-following property and therefore satisfies the following constraints:
\begin{itemize}
    \item \textbf{Sentence Preservation}: Each sentence $s_t$ is either mapped to a node $n_i$ or to a connection $c_j$;
    \item \textbf{Order Traceability}: The original sentence ordering in $\mathcal{S}$ is preserved and recorded during transformation;
    \item \textbf{Content Preservation}: The full content of $\mathcal{S}$ can be reconstructed from $(\mathcal{N}, \mathcal{C})$ in any valid topological ordering of $\mathcal{G}$.
\end{itemize}

Let $\mathcal{T}$ denote the set of all total topological orderings over the DAG $\mathcal{G}$. Then each $\mathcal{S}\tau$ is a CoT realization consistent with the dependency structure in $\mathcal{G}$, then we have:
\begin{equation}
P(\mathcal{G}) = \bigcup_{\tau \in \mathcal{T}} P(\mathcal{S}_\tau),
\end{equation}

\paragraph{Proposition.}
Let $\tau \in \mathcal{T}$ be the topological sort that matches the original sentence order of $\mathcal{S}$ (as recorded by $\phi$). Then:
\begin{equation}
\text{LLM}(\mathcal{S}) = \text{LLM}(\mathcal{S}{\tau}) \overset{\phi}{\equiv} \text{LLM}(\mathcal{G} \mid \tau).
\end{equation}
which shows that under the fixed ordering $\tau$, the original CoT and the structured DAG form are semantically equivalent, and this alignment is preserved even after pruning (as pruning respects $\tau$).

\paragraph{Proof Sketch.}
Let $\tau\!:\mathcal{N}\!\to\!\{1,\dots,T\}$ be the topological ordering that preserves the original sentence order, namely $\tau(n_i)=t$ if and only if node $n_i$ is derived from sentence $s_t$.  
For every node $n_i$ the minimal support set therefore satisfies
\begin{equation}
F_i \;=\; \bigl\{\,n_j \mid \tau(n_j)<\tau(n_i)\bigr\}
\;\longleftrightarrow\; s_{<\tau(n_i)} .
\label{eq:support}
\end{equation}

Substituting Equation~\eqref{eq:support} into the DAG factorisation gives
\begin{equation}
\begin{split}
P(\mathcal{G}\mid\tau)
&= \prod_{t=1}^{T}P\bigl(n_{\tau^{-1}(t)}\mid F_{\tau^{-1}(t)}\bigr)\\
&= \prod_{t=1}^{T}P(s_t\mid s_{<t})
= P(\mathcal{S}).
\end{split}
\label{eq:chain}
\end{equation}

Because an autoregressive LLM uses the same chain rule, we have the desired equivalence:
\begin{equation}
\operatorname{LLM}(\mathcal{S})
= \operatorname{LLM}\!\bigl(\mathcal{S}_{\tau}\bigr)
\;\overset{\phi}{\equiv}\;
\operatorname{LLM}\!\bigl(\mathcal{G}\mid\tau\bigr).
\end{equation}

Building on the previous lemmas, we now extend this argument to the pruned graph.  For each candidate node \(n_i\), define its loss increment consistent with Section 3.3:
\begin{equation}
\Delta_i
\;=\;
\log \mathcal{M}\bigl(\mathcal{S}_{\tau}\bigr)
\;-\;
\log \mathcal{M}\bigl(\mathcal{S}_{\tau}\setminus\{n_i\}\bigr),
\end{equation}
which measures the increase in negative log-probability when \(n_i\) is removed.  Rank all eligible \(n_i\) by $\lvert\Delta_i\rvert$, delete the $k \;=\;\bigl\lfloor \gamma\,\lvert\mathcal{N}\rvert\bigr\rfloor$ nodes with smallest $\lvert\Delta_i\rvert$, and for each removed node splice its unique predecessor to its unique successor with a “shortcut” edge.  Denote the resulting pruned graph by \(\mathcal{G}'=(\mathcal{N}',\mathcal{E}')\), the pruned sequence by \(\mathcal{S}_{\mathrm{pruned}}\), and let $\varepsilon \;=\;\max_{i\in\mathrm{removed}} \lvert\Delta_i\rvert.$
Since removing each node changes the log-probability of the linearized CoT by at most $\lvert\Delta_i\rvert$, and the shortcuts restore a valid autoregressive order $tau'=\tau|_{\mathcal{N}'}$, for any autoregressive model $\mathcal{M}$ we obtain
\begin{equation}
\bigl|\log\mathcal{M}(\mathcal{S})
-\log\mathcal{M}(\mathcal{S}_{\mathrm{pruned}})\bigr|
\le \sum_{i\in\mathrm{removed}}|\Delta_i|
\le k\,\varepsilon,
\end{equation}

\begin{equation}
\bigl|\log\mathcal{M}(\mathcal{G}'\mid\tau')
-\log\mathcal{M}(\mathcal{S}_{\mathrm{pruned}})\bigr|
\le k\,\varepsilon.
\end{equation}
Thus, we conclude that for any logic graph \(\mathcal{G} = (\mathcal{N},\mathcal{E})\) produced by the semantics‐preserving transformation \(\phi\) — and for its pruned variant \(\mathcal{G}' = (\mathcal{N}',\mathcal{E}')\) obtained by deleting the \(k = \lfloor \gamma\,|\mathcal{N}|\rfloor\) lowest‐utility nodes and adding shortcut edges — the linearizations \(\mathcal{S}_\tau\) and \(\mathcal{S}_{\mathrm{pruned}}\) under the original sentence order \(\tau\) yield exactly the same LLM output up to a tight log‐probability deviation of $k\,\varepsilon$, where $k = \bigl\lfloor \gamma\,|\mathcal{N}|\bigr\rfloor$, $\varepsilon = \max_{i\in\mathrm{removed}}|\Delta_i|$.

\section{Additional Experimental Results}
\label{app:additional}
\paragraph{Verification pruning generalizes beyond math reasoning.}
We further assess verification pruning on an out-of-math task—CommonsenseQA—to test its generalization to open-domain reasoning. As shown in Figure~\ref{fig:commonsenseqa}, pruning verification steps yields substantial performance gains even outside the mathematical domain. Specifically, applying the Cut-All-Verification strategy improves accuracy from 37.8\% to 54.1\% (+16.3), while simultaneously reducing token usage by 16.2\%. This indicates that verification spans often contain redundant self-reflection or verbose elaboration, which can be safely removed without sacrificing—and in fact enhancing—reasoning quality. Notably, partial pruning (e.g., Cut-0.5 or Cut-0.7) underperforms full pruning, suggesting that removing all verification logic is more robust in this task. These findings underscore that verification pruning generalizes beyond math and is a domain-agnostic strategy for balancing accuracy and efficiency in SLMs.

\paragraph{Verification pruning remains effective regardless of Long-CoT supervision.}
Table~\ref{tab:wolong} compares verification pruning on two models with similar scale (8B) but differing training regimes: \textbf{DeepSeek-R1-Distill-LLaMA-8B}, which is trained with Long-CoT traces, and \textbf{LLaMA3.1-8B-Instruct}, which lacks such supervision. Despite this difference, both models benefit from pruning. Notably, \textit{R1-Distill-LLaMA-8B} achieves up to \textbf{+15.0\% accuracy gain} (AMC23: 80.0 vs. 65.0) and \textbf{-5.7\% token reduction}, showing that even well-trained Long-CoT models contain residual redundant logic that pruning can filter for improved reasoning clarity. Conversely, \textit{LLaMA3.1-8B} benefits more in terms of efficiency—achieving \textbf{-9.1\% token compression}—despite only modest accuracy improvements. This reflects its noisier, less structured reasoning traces due to lacking CoT supervision, which verification pruning trims more aggressively. These results suggest verification pruning is broadly applicable: for CoT-tuned models, it distills the most decisive steps; for non-CoT models, it removes guesswork and improves interpretability. The pruning operation therefore acts as a general reasoning sanitizer across model capabilities.

\paragraph{Verification pruning benefits scale with model size—even for small models.}
As shown in Table~\ref{tab:1_5b}, pruning verification nodes consistently improves both efficiency and performance. On the smaller DeepSeek-R1-Distill-Qwen-1.5B, Cutting All Verification strategy raises accuracy from 34.5\% to 41.4\%—including a notable +20.0 improvement on AMC23—while reducing token usage by 1.95\%. These results indicate that verification steps encode high-utility reasoning signals, and pruning enables small models to better leverage them. For the larger 7B variant, pruning delivers even greater benefits: accuracy climbs to 63.0\% with a 9.5\% token reduction. This scaling trend suggests that larger models exhibit more verifiable redundancy, which structure-aware pruning can exploit to improve both token efficiency and accuracy simultaneously.

\begin{table*}[ht!]
\centering
\begin{tabular}{llcccccc}
\toprule
\multirow{2}{*}{\makecell{\textbf{Pruning Ratio}\\ ~$\boldsymbol{\gamma}$}}
& 
\multicolumn{3}{c}{\textbf{In-Domain}} & 
\multicolumn{2}{c}{\textbf{Out-of-Domain}} & 
\multicolumn{2}{c}{\textbf{Overall Performance}} \\
\cmidrule(lr){2-4} \cmidrule(lr){5-6} \cmidrule(lr){7-8}
& AMC23$\uparrow$ & AIME$\uparrow$ & MATH500$\uparrow$ & GSM8K$\uparrow$ & BBH$\uparrow$ & Avg. Acc.$\uparrow$ & Token ($\Delta$)$\uparrow$ \\
\midrule
\rowcolor{gray!8}
\multicolumn{8}{c}{DeepSeek-R1-Distill-Llama-8B~\citep{guo2025deepseek}} \\
\midrule
Original & 65.0 & 25.0 & 81.2 & 80.4 & 37.0 & \underline{57.7} & 4555.9 \\
\midrule
\multicolumn{8}{l}{{\textit{Pruning Nodes~over Verification}}} \\
\;\; Cut 0.5 & 57.5  & 26.7 & \underline{83.2} &    80.0   &   35.2    & 56.5 & {4786.2 (+5.05\%)} \\
\;\; Cut 0.7 & 60.0  & 28.3 & 82.8 &    79.0   &   35.2    & 57.1 & {4392.6 (-3.58\%)} \\
\;\; Cut 0.9 &   37.5    & \underline{30.0} & 82.2 &     81.4  &    37.1   & 53.6 & {4428.3 (-2.80\%)} \\
\rowcolor[RGB]{236,244,252}
\;\; Cut All & \textbf{80.0}  & \underline{30.0} & \textbf{84.2} & \textbf{82.5} &   \textbf{38.0}   & \textbf{62.9} & \textbf{4295.2 (-5.72\%)} \\
\midrule 
\rowcolor{gray!8}
\multicolumn{8}{c}{Llama3.1-8B-Instruct~\cite{grattafiori2024Llama}} \\
\midrule
Original & 22.5 & 1.7 & \textbf{43.4} & \textbf{83.5} & 9.3 & 32.1 & 3488.5 \\
\midrule
\multicolumn{8}{l}{{\textit{Pruning Nodes~over Verification}}} \\

\;\; Cut 0.5 & 5.0 & 0.0 & 40.4 & 58.8 & 11.1 & 23.1 & {3505.7 (+0.49\%)} \\
\;\; Cut 0.7 & 10.0 & 1.7 & 40.6 & 55.7 & 15.7 & 24.7 & {3489.8 (+0.04\%)}  \\
\;\; Cut 0.9 & 25.0 & 1.7 & 41.8 & 62.9 & 15.7 & 29.4 & {3584.6 (+2.75\%)}  \\
\rowcolor[RGB]{236,244,252}
\;\; Cut All & \textbf{30.0}  & \textbf{1.7} & 38.6 & \textbf{83.5} &   \textbf{15.7}   & \textbf{33.9} & \textbf{3171.1 (-9.10\%)} \\

\bottomrule
\end{tabular}
\caption{
Verification pruning improves reasoning in both Long-CoT-capable and non-capable models.
}
\label{tab:wolong}
\end{table*}

\begin{table*}[ht!]
\centering

\begin{tabular}{llcccccc}
\toprule
\multirow{2}{*}{\makecell{\textbf{Pruning Ratio}\\ ~$\boldsymbol{\gamma}$}}
& 
\multicolumn{3}{c}{\textbf{In-Domain}} & 
\multicolumn{2}{c}{\textbf{Out-of-Domain}} & 
\multicolumn{2}{c}{\textbf{Overall Performance}} \\
\cmidrule(lr){2-4} \cmidrule(lr){5-6} \cmidrule(lr){7-8}
& AMC23$\uparrow$ & AIME$\uparrow$ & MATH500$\uparrow$ & GSM8K$\uparrow$ & BBH$\uparrow$ & Avg. Acc.$\uparrow$ & Token ($\Delta$)$\uparrow$ \\

\midrule 
\rowcolor{gray!8}
\multicolumn{8}{c}{DeepSeek-R1-Distill-Qwen-1.5B~\cite{guo2025deepseek}} \\
\midrule
Original & 22.5 & 16.7 & 60.4 & 60.8 & 11.9 & 34.5 & 7996.5 \\
\midrule

\multicolumn{8}{l}{{\textit{Pruning Nodes~over Verification}}} \\
\;\; Cut 0.5 & 22.5  & \textbf{23.3} & 57.8 &    59.8   &   \textbf{16.7}    & 36.0 & {8237.0 (+3.01\%)} \\
\;\; Cut 0.7 & \underline{32.5}  & 16.7 & 60.4 &    \underline{62.9}   &   10.2    & \underline{36.5} & {8392.2 (+4.95\%)} \\
\;\; Cut 0.9 &   15.0    & \textbf{23.3} & \underline{61.4} &     56.7  &    10.2   & 33.3 & {8218.6 (+2.78\%)} \\
\rowcolor[RGB]{236,244,252}
\;\; Cut All & \textbf{42.5}  & \underline{20.0} & \textbf{61.6} & \textbf{71.1} &   \underline{12.0}   & \textbf{41.4} & \textbf{7840.9 (-1.95\%)} \\

\midrule
\rowcolor{gray!8}
\multicolumn{8}{c}{DeepSeek-R1-Distill-Qwen-7B~\cite{guo2025deepseek}} \\
\midrule
Original & 60.0 & 33.3 & 82.8 & 88.6 & 20.4 & 57.0 & 4344.3 \\
\midrule
\multicolumn{8}{l}{{\textit{Pruning Nodes~over Verification}}} \\
\;\; Cut 0.5 &   57.5    & 33.4 &   83.8  &     \textbf{89.7}  &  22.2 & 57.3 & {3998.7 (-7.96\%)} \\
\;\; Cut 0.7 &   55.0    & \underline{41.7} & 83.0 &     \textbf{89.7}  &  21.3    &  58.1 & {4169.0 (-4.03\%)}  \\
\;\; Cut 0.9 & 65.0  & 
36.7 & \underline{86.0} &    88.7   &    18.5   &  \underline{59.0} &  {4258.5 (-1.97\%)} \\
\rowcolor[RGB]{236,244,252}
\;\; Cut All & \textbf{72.5} & \textbf{43.3} & \underline{86.0} & \underline{89.0} &  \textbf{25.9} & \textbf{63.0} & \textbf{3931.2 (-9.51\%)}  \\
\bottomrule
\end{tabular}
\caption{
Verification-based pruning enhances token efficiency with minimal accuracy trade-offs, especially for larger models.
}
\label{tab:1_5b}
\end{table*}

\section{Case Study}
This section presents two illustrative examples to analyze our method qualitatively. First, we showcase a pruned training data example derived from logic graph segmentation (
\ref{subapp:data}). Second, we analyze the inference behavior of models fine-tuned using our Long2Short CoT supervision(\ref{subapp:case}).

\subsection{An example of Pruning-on-Logic Data}
\label{subapp:data}
Table~\ref{tab:qa-token-table} presents a randomly selected training data example comparing the original Long-CoT reasoning with our pruned version based on logic graph segmentation. This illustrates how pruning eliminates redundant narrative while retaining essential computation and reasoning steps.

\subsection{Analysis of Verification Pruning in SLMs}
\label{subapp:case}
Verification pruning surprisingly preserves semantic coherence while selectively removing non-essential verification steps. As shown in Tables~\ref{tab:strategies_comparation1} and \ref{tab:strategies_comparation2}, pruning verification nodes eliminates redundant self-checking without disrupting the reasoning structure, yielding the correct answer by focusing on core computations (e.g., $\binom{5}{4}$ and $\binom{10}{8}$). This observation is consistent with our main results in Section~4.2, which demonstrate that verification segments contribute minimally to reasoning quality, whereas indiscriminate chain- or reasoning-level pruning disrupts essential logical dependencies. These findings confirm that pruning can effectively identify and remove dispensable parts of reasoning traces, improving efficiency without compromising semantic integrity.

\begin{table*}[htbp]
\centering
\begin{tabular}{@{} p{0.3\textwidth} p{0.5\textwidth} c @{}}
\toprule
\textbf{Type} & \textbf{Content} & \textbf{Token Count} \\
\midrule
Question & A room is 19 feet long and 11 feet wide. Find the ratio of the length of the room to its perimeter. Express your answer in the form $a:b$. & - \\
\textbf{LongCoT (Original)} &
\begin{minipage}[t]{8cm}
\textless{}\textbar{}begin\_of\_thought\textbar{}\textgreater{}Okay, so I need to find the ratio of the length of a room to its perimeter. The room is 19 feet long and 11 feet wide. Let me start by recalling what perimeter means. The perimeter of a rectangle is calculated by adding up all the sides, right? Since a rectangle has two lengths and two widths, the formula should be P = 2*(length + width). Let me check that. Yeah, that makes sense because you're doubling the sum of one length and one width to account for both pairs of sides. First, let me write down the given dimensions. The length is 19 feet, and the width is 11 feet. So plugging those into the perimeter formula, the perimeter P would be 2*(19 + 11). Let me compute that. 19 plus 11 is 30, and then multiplying by 2 gives 60. So the perimeter is 60 feet. Now, the problem is asking for the ratio of the length to the perimeter. The length is 19 feet, and the perimeter is 60 feet. So the ratio would be 19:60. But wait, I should make sure that this ratio is in its simplest form. To do that, I need to find the greatest common divisor (GCD) of 19 and 60 and divide both numbers by it. Let me think about the factors of 19. Since 19 is a prime number, its only factors are 1 and 19. The factors of 60 are 1, 2, 3, 4, 5, 6, 10, 12, 15, 20, 30, 60. The common factors of 19 and 60 are just 1. That means the ratio 19:60 can't be simplified further. So the ratio is already in its simplest form. Let me double-check my calculations to be sure. The perimeter: 2*(19+11) = 2*30 = 60. Correct. The length is 19. So ratio 19:60, which simplifies to 19:60 since they have no common factors other than 1. Yep, that seems right. So the final answer should be 19:60.  \textless{}\textbar{}end\_of\_thought\textbar{}\textgreater{}\\
\end{minipage}
& 503 \\
\textbf{Prune-on-Logic (CutallNode Verification)} & \begin{minipage}[t]{8cm}
\textless{}\textbar{}begin\_of\_thought\textbar{}\textgreater{}Okay, so I need to find the ratio of the length of a room to its perimeter.  
The room is 19 feet long and 11 feet wide.  

Since a rectangle has two lengths and two widths, the formula should be P = 2*(length + width).  

The length is 19 feet, and the width is 11 feet.  
So plugging those into the perimeter formula, the perimeter P would be 2*(19 + 11).  
19 plus 11 is 30, and then multiplying by 2 gives 60.  
So the perimeter is 60 feet.  

Now, the problem is asking for the ratio of the length to the perimeter.  
The length is 19 feet, and the perimeter is 60 feet.  
So the ratio would be 19:60.\textless{}\textbar{}end\_of\_thought\textbar{}\textgreater{}
\end{minipage}
 & 181 \\
\bottomrule
\end{tabular}
\caption{Comparison of Question, LongCoT, and Prune-on-Logic Outputs with Token Counts.}
\label{tab:qa-token-table}
\end{table*}


\begin{table*}
\begin{tabular}{@{} p{0.3\textwidth} p{0.5\textwidth} c @{}}
\toprule
\textbf{Type} & \textbf{Content} & \textbf{Token Count} \\
\midrule
Question & The Greek army contained two types of soldiers: the upper class and the lower class soldiers. If there were a total of 5 upper class soldiers, and 10 lower class soldiers in a certain part of Athens, and the battle of Thermopylae demands a force of 4 upper class soldiers and 8 lower class soldiers, how many different battalions can be sent? (correct answer: $\boxed{225}$) & - \\
\textbf{Original} &
\begin{minipage}[t]{0.5\textwidth}
\textless{}think\textgreater{}Okay, so I need to figure out how many different battalions can be sent for the battle of Thermopylae given the number of upper and lower class soldiers available ... \\
Final Answer:The number of different battalions that can be sent is \\boxed{1}.\textless{}/think\textgreater{}...Both calculations show that the maximum number of battalions that can be formed is 1, as the limiting factor is the number of upper class soldiers. This conclusion holds even when considering that each soldier can only be used once in a single battalion.Thus, the number of different battalions that can be sent is $\boxed{1}$.
\end{minipage}
& 3529 \\
\textbf{Cut all nodes over chain} & \begin{minipage}[t]{0.5\textwidth}
\textless{}think\textgreater{}Okay, so I have this problem about the Greek army with two types of soldiers: upper class and lower class. The question is asking how many different battalions can be sent to meet the demand for the battle of Thermopylae. Let me try to break it down step by step  ... \\
Wait, 5 choose 4 is actually 5 because you're choosing 4 out of 5, which leaves only one soldier not chosen. So, yeah, that's correct.For C(10,8), that's equivalent to C(10,2) because choosing 8 out of 10 is the same as leaving out 2. So,C(10,2) = 10! / (2! * (10-2)! ) = (10*9)/2 = 45.Alternatively, calculating directly:C(10,8) = 10! / (8! * 2!) = (10*9)/2 = 45. Yep, same result. 

\end{minipage}
 & 512 \\

\bottomrule
\end{tabular}
\caption{Inference Analysis of SLM Tuned with Long2Short (continued on next page).}
\label{tab:strategies_comparation1}
\end{table*}

\onecolumn
\begin{table*}[htbp]
\centering
\begin{tabular}{@{} p{0.3\textwidth} p{0.5\textwidth} c @{}}
\toprule
\textbf{Type} & \textbf{Content} & \textbf{Token Count} \\
\midrule

\textbf{Cut all nodes over resoning} & \begin{minipage}[t]{8cm}
\textless{}think\textgreater{}Okay, so I have this problem about the Greek army with two types of soldiers: upper class and lower class. The numbers given are 5 upper class soldiers and 10 lower class soldiers in a certain part of Athens. The battle of Thermopylae requires 4 upper class soldiers and 8 lower class soldiers. The question is asking how many different battalions can be sent. Hmm, let me try to figure this out step by step ... \\
Therefore, the answer should be 6 different battalions.**Final Answer**The number of different battalions that can be sent is \boxed{6}.\textless{}/think\textgreater{}To determine how many different battalions can be sent to the battle of Thermopylae, we need to consider the constraints on the number of upper class (U) and lower class (L) soldiers required (4U and 8L) and the available numbers (5U and 10L)...Listing all combinations confirms there are 6 different battalions:- (4, 8), (4, 9), (4, 10)- (5, 8), (5, 9), (5, 10)Thus, the number of different battalions that can be sent is \(\boxed{6}\).

\end{minipage}
 & 925 \\
 
 \textbf{Cut all nodes over verification} & \begin{minipage}[t]{8cm}
\textless{}think\textgreater{}Okay, so I need to figure out how many different battalions can be sent for the battle of Thermopylae given the number of upper and lower class soldiers available. Let me start by understanding the problem again.There are two types of soldiers: upper class and lower class. In a certain part of Athens, there are 5 upper class soldiers and 10 lower class soldiers. The battle requires 4 upper class soldiers and 8 lower class soldiers... \\
**Final Answer**The number of different battalions that can be sent is \boxed{225}.\textless{}/think\textgreater{}...First, we calculate the number of ways to choose 4 upper class soldiers from 5:\[\binom{5}{4} = \frac{5!}{4!(5-4)!} = \frac{5 \times 4!}{4! \times 1!} = 5\]Next, we calculate the number of ways to choose 8 lower class soldiers from 10:\[\binom{10}{8} = \frac{10!}{8!(10-8)!} = \frac{10 \times 9 \times 8!}{8! \times 2!} = \frac{10 \times 9}{2 \times 1} = 45\]Multiplying these two results gives the total number of different battalions:\[5 \times 45 = 225\]Thus, the number of different battalions that can be sent is \(\boxed{225}\).

\end{minipage}
 & 1095 \\

\bottomrule
\end{tabular}
\caption{Inference Analysis of SLM Tuned with Long2Short (continued).}
\label{tab:strategies_comparation2}
\end{table*}

\end{document}